# Temporal Binding Foundation Model for Material Property Recognition via Tactile Sequence Perception

Hengxu You[1], Tianyu Zhou[1], and Jing Du[1,2*]

[1] Department of Civil & Coastal Engineering, University of Florida, Gainesville, FL 32611, USA
[2] Department of Mechanical & Aerospace Engineering, University of Florida, Gainesville, FL 32611, USA
* Corresponding Author

**Abstract**—Robots engaged in complex manipulation tasks require robust material property recognition to ensure adaptability and precision. Traditionally, visual data has been the primary source for object perception; however, it often proves insufficient in scenarios where visibility is obstructed or detailed observation is needed. This gap highlights the necessity of tactile sensing as a complementary or primary input for material recognition. Tactile data becomes particularly essential in contact-rich, small-scale manipulations where subtle deformations and surface interactions cannot be accurately captured by vision alone. This letter presents a novel approach leveraging a temporal binding foundation model for tactile sequence understanding to enhance material property recognition. By processing tactile sensor data with a temporal focus, the proposed system captures the sequential nature of tactile interactions, similar to human fingertip perception. Additionally, this letter demonstrates that, through tailored and specific design, the foundation model can more effectively capture temporal information embedded in tactile sequences, advancing material property understanding. Experimental results validate the model's capability to capture these temporal patterns, confirming its utility for material property recognition in visually restricted scenarios. This work underscores the necessity of embedding advanced tactile data processing frameworks within robotic systems to achieve truly embodied and responsive manipulation capabilities.

**Index Terms**—Tactile Sensing, Material Property Recognition, Foundation Model, Robotic Perception

## I. INTRODUCTION

The ability of robots to recognize and adapt to different material properties is a crucial component in developing versatile and intelligent manipulation systems [1-4]. Traditionally, visual sensors have been the primary means of perceiving and identifying objects, allowing robots to make informed decisions based on texture, shape and spatial orientation [5, 6]. However, reliance solely on visual data can be limiting, particularly in scenarios where visibility is obstructed or when detailed tactile information is needed to distinguish between materials with similar visual characteristics. This gap underscores the importance of tactile sensing as a complementary modality, enabling robots to physically interact with their environment in a manner akin to human touch [7, 8].

Tactile sensing is especially valuable in contact-rich, small-scale manipulation tasks, such as handling soft or deformable materials. Studies have shown that tactile sensors, when integrated with machine learning models, can significantly enhance a robot's perception and handling capabilities by providing detailed information about texture, hardness, and compliance [9-12]. Tactile data also supports adaptive force feedback control, enabling robots to dynamically adjust their grip and manipulation strategy based on the material's properties, thus contributing to safer and more effective interactions [13, 14].

Recent advancements in tactile sensor design have led to the development of high-resolution fingertip sensors that push robotic tactile sensing closer to human-level sensory capabilities [15-17]. These sensors capture fine-grained details, allowing robots to perceive subtle differences in texture and pressure distribution with improved accuracy. However, interpreting this high-resolution tactile input as a new sensory modality presents unique challenges, requiring models capable of handling complex, data-rich inputs. Foundation models, with their strong transferability and adaptability, provide an effective approach for analyzing tactile data, capturing detailed and temporally correlated patterns essential for comprehensive material recognition [18, 19]. The necessity of foundation models in tactile-based learning has become increasingly evident as robots acquire more sophisticated and data-intensive sensory inputs [9, 20].

Furthermore, human tactile perception is inherently temporal and incorporates memory mechanisms that facilitate continuous learning and adaptation. As noted in [21], human touch relies on past sensory experiences to inform current responses. This insight underscores the value of incorporating similar temporal learning mechanisms in robotic tactile systems to achieve human-like processing. By

Corresponding author: Eric Du (eric.du@essie.ufl.edu). IEEE Sensors Letters discourages courtesy authorship; please use the Acknowledgment section to thank your colleagues for routine contributions.

Digital Object Identifier: 10.1109/LSEN.XXXX.XXXXXXX (inserted by IEEE).

embedding memory-based structures into tactile learning models, robots can better replicate human sensory behaviors, leading to more adaptive and precise material property recognition. Advanced architectures, such as those utilizing transformers and temporal networks, provide a robust framework for processing sequential high-resolution data [22].

Inspired by human tactile memory and supported by the development of foundation model on perception, this letter presents a temporal binding foundation model, TeBi-Llama (Llama model with temporal information binding), that processes tactile data as time-sequenced input. Our model captures the nuanced temporal patterns embedded in tactile interactions, enhancing the robot's ability to recognize and adapt to different material properties.

## II. METHODS

### A. System Design

Our proposed TeBi-Llama model is based on the model proposed in [20], which applies LlaMa-Adapter and ImageBind to align the tactile and RGBD as multi-modality inputs with LlaMa2 model. On top of the original model, we designed a LSTM-based encoder to process the sequential image-tactile input pairs. To effectively excavate the embedded information in the sequential sensory inputs, we develop a structure-aware cross-modality feature fusion method that gradually insert the hidden states into different stages of the attention layers in LlaMa2. Fig. 1 gives the overall structure of our designed model including the two major modules: **LSTM Encoder** and **Structure-Aware Temporal Fusion**.

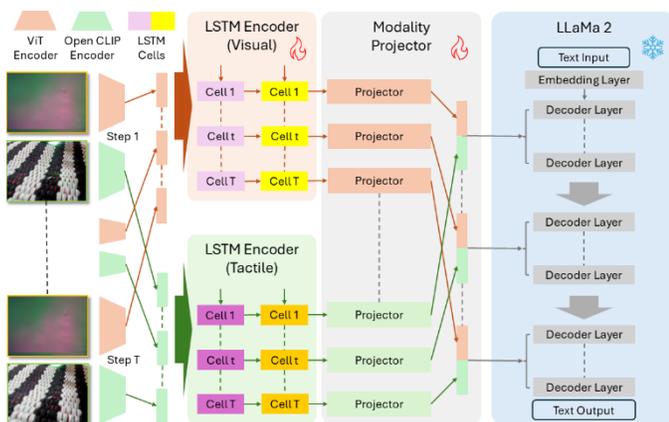

Fig. 1. Resonance frequency as a function of time. Note that "Fig." is abbreviated. It is good practice to explain the significance of the figure in the caption.

### B. LSTM Encoder

Fig. 2 gives the detailed design of the LSTM encoder. Let $T$ to be the sequence length. $Im^t$ and $Tac^t$ denote the image and tactile frame at time step $t$. We apply the pretrained visual encoder from OpenCLIP [23] as the image encoder and the Vision Transformer [24] (ViT) as the tactile encoder, which align with the usage in [20]. The encoded feature pairs ($f_{image}^t, f_{tactile}^t$) would be separately fed to two LSTM modules with similar structures. Let $h_{image}^t$ and $h_{tactile}^t$ denote the hidden state output at $t$ for image and tactile respectively.

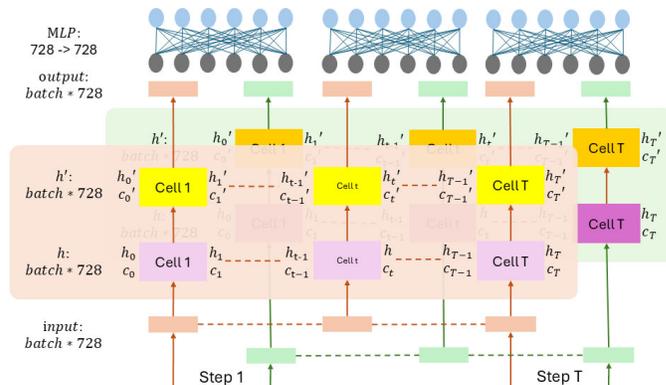

Fig. 2. Resonance frequency as a function of time. Note that "Fig." is abbreviated. It is good practice to explain the significance of the figure in the caption.

### C. Structure-aware Temporal Fusion

We design a structure-aware feature fusion method that insert the temporal features from LSTM into different attention layers in the foundation model illustrated in Fig. 3. We apply the opensource Llama2-7b as the main model.

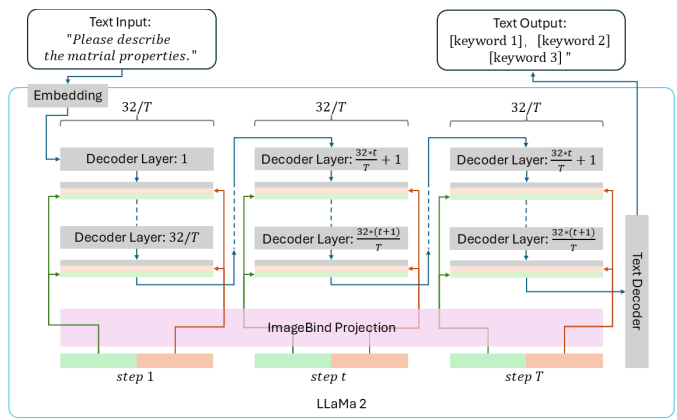

Fig. 3. Resonance frequency as a function of time. Note that "Fig." is abbreviated. It is good practice to explain the significance of the figure in the caption.

Instead of using the last-step LSTM output as Llama2 input once, we utilized each hidden to capture the evolution of the tactile-visual sequence, as each state contains valuable intermediate information that contributes to the final estimation. We apply ImageBind [25] to fuse the LSTM-generated hidden states with the LLaMA-2 model's attention layers. Given that LLaMA-2 consists of $n$ attention layers, we map each hidden state outputs $h^t := <h_{image}^t, h_{tactile}^t>$ to a distinct subset of the attention layers. Assuming $T$ dived evenly into 32, we assign each $h^t$ to a set of $32/T$ consecutive attention layers. Specifically, for the $32/T$, we fuse $h^1$ (separately on $h_{image}^1$ and $h_{tactile}^1$) with the output of each layer; for the next $32/T$ layers, we fuse $h^2$; continuing this process until the last $32/T$ layers are fused with $h^T$. This design aligns each LSTM hidden state with the LLaMA-2 layers, reflecting a gradual refinement of the multi-modal features.

Aligning each LSTM hidden state with successive blocks of LLaMA-2's attention layers leverage its hierarchical processing structure. Visual-tactile input, with its changing sensory details, benefits from this approach. Early LSTM states, which tends to capture transient or noisy information, pair with the initial LLaMA-2 layers that focus on broad features. As the sequence progresses, more refined LSTM states align with deeper LLaMA-2 layers, which interpret high-level patterns like consistent textures. This structured flow enables clearer, interpretable multi-modal feature extraction. The progressive alignment from to across the 32 layers supports the sequential nature of visual-tactile data, enhancing the model's ability to track both short-term variations and long-term dependencies crucial for manipulation tasks.

## III. RESULTS AND ANALYSIS

### A. Dataset

The training dataset is Human Collected Tactile (HCT) Dataset from [20]. The HCT dataset consists of 39,154 visual-tactile image pairs, collected from five human subjects across 847 trajectories (a total of 20 hours). Each trajectory captures four stages of interaction: approaching, contacting, sliding, and withdrawing, where participants use a GelSight sensor to interact with various objects.

To align with the LSTM training setup, we reorganized the original dataset to create the HCT Temporal training set. For each trajectory, we sampled frames starting from the first frame to the $T$-th frame from the end, where is the chosen sequence length. Given a trajectory length of $L_{traj}$, this setup yields $L_{traj} - T + 1$ temporal training examples per trajectory. Each example consists of $T$ consecutive tactile-RGB pairs, paired with a text description of the object's material properties (e.g., "solid," "soft"). Since only one object is used in each trajectory, the material description remains consistent across all frames in a single LSTM sample $Tem^i$ as:

$$Tem^i := <[Im^t, Tac^t],[keyword\ labels]>, t \in [1, T] \quad (1)$$

### B. Training Setup

We employed a two-stage training workflow consisting of LSTM encoder pretraining and temporal-binding finetuning, following a setup similar to that in [20].

In the pretraining phase, the task was framed as a 402-class classification problem, using the text description of each data sample as the ground truth. We evaluated model performance through top-1 and top-5 accuracy metrics for both tactile-text and tactile-visual modalities. The visual-text group was omitted, as it relies on a pre-trained feature extractor from CLIP, which is not directly relevant to the current model's design analysis. Here, top-1 accuracy represents instances where the ground truth label matches the model's highest-confidence prediction, while top-5 accuracy reflects cases where the label appears within the top five predictions by confidence. The LSTM encoder for both tactile and visual modalities consist of two unidirectional LSTM layers, followed by two MLP layers to align the feature output size with the LlaMa2 input requirements.

In finetuning phase, we applied the proposed temporal binding method, integrating the pre-trained LSTM encoder features into the LLaMA-2 7B model's 32 multi-head self-attention layers. This approach enables the progressive alignment of sequential features from the LSTM encoder with LLaMA-2's hierarchical processing layers.

The training process used 8 A100 GPUs and 96 CPU cores on the University of Florida's HiPerGator computing platform, supported by NVIDIA. This setup is used to train both the LSTM encoder and the binding model, ensuring efficient utilization of computational resources.

### C. Results

To illustrate the improvements achieved by applying the temporal binding structure, we compared results in two areas: the LSTM encoder classification results after pretraining and the overall model performance on material property recognition.

Table 1. Top-1 and top-5 accuracies comparison.

| Model | Top1 Accuracy | Top5 Accuracy |
|---|---|---|
| Base | 51.12 | 74.41 |
| LSTM | 62.75 | 91.37 |

Table. 1 presents the top-1 and top-5 accuracy metrics for the LSTM encoder following pretraining. For comparison, we included results from the baseline model, which utilizes a single-frame tactile and image encoder. To maintain consistency, we selected only $Im^T$ and $Tac^T$ from each LSTM sample to form the training set for the baseline model. Results indicate that the LSTM encoder outperforms the baseline in both top-1 and top-5 accuracy.

We observed a more significant increase in top-1 accuracy than in top-5 accuracy, suggesting that while both the baseline and LSTM encoders provide a reasonably accurate range of material properties (as seen in top-5 accuracy), the LSTM encoder offers more precise recognition capabilities. This indicates that the LSTM encoder is better at accurately identifying the correct material properties within the top prediction, demonstrating the benefit of temporal feature integration for material property recognition.

Table 2. Overall model performances comparison.

| Model | Group | Evaluation Score (0 - 5) |
|---|---|---|
| Base Model | Tactile & vision | 3.736 |
|  | Tacti | 1.414 |
| LSTM Even | Tactile & vision | 3.582 |
|  | Tactile only | 2.480 |
| LSTM Aware | Tactile & vision | 4.031 |
|  | Tactile only | 2.605 |

Table 2 presents the test results of the overall model's performance on material property recognition. We applied a 10% train-test split and evaluated three configurations: the Base Model with single-frame input, the LSTM encoder without temporal and structural awareness (LSTM Even), and the LSTM encoder with both temporal and structural awareness (LSTM Aware). For the LSTM Even model, we extracted only the last hidden state output from the LSTM encoder and fused it across all attention layers. Following a method similar to [20], we used ChatGPT4-v to score the similarity between the model's predictions and ground truth, with scores ranging from 0 to 5.

In the tactile-vision group, we observed that the LSTM Even model actually scored lower than the Base Model, indicating that simply using the last hidden state without considering the structure of the attention layers diminished performance. In contrast, the LSTM Aware model showed significant improvement over the Base Model, demonstrating that a structured fusion approach, which aligns the attention layer depths with the temporal information flow, is crucial for optimal performance.

In the tactile-only group, we saw a continuous improvement in scores from the Base Model to LSTM Even and further to LSTM Aware. The Base Model achieved a score of 1.414, which is significantly lower than the scores observed in the tactile-vision group. This highlights the challenges of using tactile data alone for material property recognition. However, with the incorporation of LSTM processing and the structured feature binding in LSTM Aware, the performance improves substantially, reaching a score of 2.605. This progression demonstrates the effectiveness of temporal modeling and structured fusion in leveraging tactile input, even in the absence of complementary visual data. The improvement from LSTM Even to LSTM Aware further demonstrates the necessity of aligning temporal features with the hierarchical structure of the foundation model to fully leverage the unique properties of tactile data.

Our results demonstrate that tactile data has strong potential to offer sensory details for material property recognition. However, a model must be specifically designed to handle the temporal flow and structured fusion of tactile features.

## IV. CONCLUSION

In this work, we explored the potential of tactile data for material property recognition, presenting a model that combines LSTM-based sequential processing with structure-aware feature fusion into the LLaMA-2 attention layers. Our approach leverages both temporal dynamics and hierarchical alignment within the model, resulting in significant improvements in recognizing material properties over baseline methods. The results demonstrate that tactile data, with its ability to capture unique sensory information, can substantially enhance robotic perception when integrated with models specifically designed to handle its temporal and structural complexities.

For future work, we plan to investigate the influence of different time steps on model performance to optimize temporal sensitivity. Additionally, recognizing that transitions between contact and non-contact frames may further reflect material properties, we aim to include integrated sequences of both frame types in training to capture these dynamic properties. We also intend to experiment with alternative binding methods, such as cross-attention, to deepen our exploration of temporal-structural fusion mechanisms.

## ACKNOWLEDGMENTS

This work was supported in part by the Nvidia AI Technology Center (NVAITC 00133684) and National Science Foundation (NSF 2024784). Any opinions, findings, and conclusions or recommendations expressed in this material are those of the authors and do not necessarily reflect the views of the NVAITC or NSF.

## REFERENCES

[1] A. Kruzliak *et al.*, "Interactive Learning of Physical Object Properties Through Robot Manipulation and Database of Object Measurements," *arXiv preprint arXiv:2404.07344*, 2024.

[2] N. Hanson, T. Kelestemur, D. Erdogmus, and T. Padir, "Pregrasp object material classification by a novel gripper design with integrated spectroscopy," *arXiv preprint arXiv:2207.00942*, 2022.

[3] V. Dave, F. Lygerakis, and E. Rückert, "Multimodal Visual-Tactile Representation Learning through Self-Supervised Contrastive Pre-Training," in *Proceedings/IEEE International Conference on Robotics and Automation*, 2024: Institute of Electrical and Electronics Engineers.

[4] H. Qi *et al.*, "General in-hand object rotation with vision and touch," in *Conference on Robot Learning*, 2023: PMLR, pp. 2549-2564.

[5] B. Zitkovich *et al.*, "Rt-2: Vision-language-action models transfer web knowledge to robotic control," in *Conference on Robot Learning*, 2023: PMLR, pp. 2165-2183.

[6] A. O'Neill *et al.*, "Open x-embodiment: Robotic learning datasets and rt-x models: Open x-embodiment collaboration 0," in *2024 IEEE International Conference on Robotics and Automation (ICRA)*, 2024: IEEE, pp. 6892-6903.

[7] H. P. Saal, I. Birznieks, and R. S. Johansson, "Memory at your fingertips: how viscoelasticity affects tactile neuron signaling," *eLife*, vol. 12, 2023.

[8] R. S. Dahiya, G. Metta, M. Valle, and G. Sandini, "Tactile sensing—from humans to humanoids," *IEEE transactions on robotics*, vol. 26, no. 1, pp. 1-20, 2009.

[9] F. Yang *et al.*, "Binding touch to everything: Learning unified multimodal tactile representations," in *Proceedings of the IEEE/CVF Conference on Computer Vision and Pattern Recognition*, 2024, pp. 26340-26353.

[10] X. Zhang *et al.*, "Target classification method of tactile perception data with deep learning," *Entropy*, vol. 23, no. 11, p. 1537, 2021.

[11] C. Higuera *et al.*, "Sparsh: Self-supervised touch representations for vision-based tactile sensing," in *8th Annual Conference on Robot Learning*.

[12] R. Calandra *et al.*, "The Feeling of Success: Does Touch Sensing Help Predict Grasp Outcomes?," in *Conference on Robot Learning*, 2017: PMLR, pp. 314-323.

[13] T. Wu, Z. Liu, Z. Ma, B. Wang, D. Ma, and H. Yu, "Bionic soft robotic gripper with feedback control for adaptive grasping and capturing applications," *Frontiers of Mechanical Engineering*, vol. 19, no. 1, pp. 1-20, 2024.

[14] D. Tian, X. Lin, and Y. Sun, "Adaptive Motion Planning for Multi-fingered Functional Grasp via Force Feedback," *CoRR*, 2024.

[15] M. Lambeta *et al.*, "Digit: A novel design for a low-cost compact high-resolution tactile sensor with application to in-hand manipulation," *IEEE Robotics and Automation Letters*, vol. 5, no. 3, pp. 3838-3845, 2020.

[16] A. Padmanabha, F. Ebert, S. Tian, R. Calandra, C. Finn, and S. Levine, "Omnitact: A multi-directional high-resolution touch sensor," in *2020 IEEE International Conference on Robotics and Automation (ICRA)*, 2020: IEEE, pp. 618-624.

[17] E. Donlon, S. Dong, M. Liu, J. Li, E. Adelson, and A. Rodriguez, "Gelslim: A high-resolution, compact, robust, and calibrated tactile-sensing finger," in *2018 IEEE/RSJ International Conference on Intelligent Robots and Systems (IROS)*, 2018: IEEE, pp. 1927-1934.

[18] R. Zhang *et al.*, "Llama-adapter: Efficient fine-tuning of language models with zero-init attention," *arXiv preprint arXiv:2303.16199*, 2023.

[19] N. Shvetsova *et al.*, "Everything at Once–Multi-modal Fusion Transformer for Video Retrieval," in *2022 IEEE/CVF Conference on Computer Vision and Pattern Recognition (CVPR)*, 2022: IEEE Computer Society, pp. 19988-19997.

[20] L. Fu *et al.*, "A Touch, Vision, and Language Dataset for Multimodal Alignment," in *Forty-first International Conference on Machine Learning*.

[21] R. S. Johansson and J. R. Flanagan, "Coding and use of tactile signals from the fingertips in object manipulation tasks," *Nature Reviews Neuroscience*, vol. 10, no. 5, pp. 345-359, 2009.

[22] B. Lim, S. Ö. Arık, N. Loeff, and T. Pfister, "Temporal fusion transformers for interpretable multi-horizon time series forecasting," *International Journal of Forecasting*, vol. 37, no. 4, pp. 1748-1764, 2021.

[23] M. Cherti *et al.*, "Reproducible scaling laws for contrastive language-image learning," in *Proceedings of the IEEE/CVF Conference on Computer Vision and Pattern Recognition*, 2023, pp. 2818-2829.

[24] A. Dosovitskiy *et al.*, "An Image is Worth 16x16 Words: Transformers for Image Recognition at Scale," in *International Conference on Learning Representations*, 2020.

[25] R. Girdhar *et al.*, "Imagebind: One embedding space to bind them all," in *Proceedings of the IEEE/CVF Conference on Computer Vision and Pattern Recognition*, 2023, pp. 15180-15190.